# Deep learning motion correction of quantitative stress perfusion cardiovascular magnetic resonance

Noortje I.P. Schueler [a], Nathan C. K. Wong [b], Richard J. Crawley [b], Josien P.W. Pluim [a], Amedeo Chiribiri [b], Cian M. Scannell [a,b]

[a] Department of Biomedical Engineering, Eindhoven University of Technology, Eindhoven, the Netherlands.
[b] School of Biomedical Engineering & Imaging Sciences, King's College London, London, United Kingdom.

**Address for correspondence:**

Dr. Cian M. Scannell

Department of Biomedical Engineering

Eindhoven University of Technology

Vector, Dominee Theodor Fliednerstraat 2,

5631 BN Eindhoven, Netherlands

Email: c.m.scannell@tue.nl

# Abstract

**Background**
Quantitative stress perfusion cardiovascular magnetic resonance (CMR) is a valuable tool for assessing myocardial ischemia. Motion correction is a crucial step in automated quantification pipelines, especially for high-resolution pixel-wise mapping. Established methods for motion correction, based on image registration, are computationally intensive and sensitive to changes in image acquisitions, necessitating more efficient and robust solutions.

**Methods**
This study developed and evaluated an unsupervised deep learning-based motion correction pipeline. Based on a previously described approach, it corrects for motion in three steps while using (robust) principal component analysis to mitigate the effects of the dynamic contrast. The time-consuming iterative registration optimizations are replaced with an efficient one-shot estimation by trained deep learning models. The pipeline aligns the perfusion series and includes auxiliary images series: the low-resolution, short-saturation preparation time arterial input function series and the proton density-weighted images. The deep learning models were trained and validated on multivendor data from 201 patients, with 38 held out for independent testing. The performance was evaluated in terms of the temporal alignment of the image series and the derived quantitative perfusion values in comparison to a previously established optimization-based registration approach.

**Results**
The deep learning approach significantly improved temporal smoothness of time-intensity curves compared to the previously published baseline (p<0.001). Temporal alignment of the myocardium (based on automated segmentations) was similar between methods and significantly improved for both as compared to before registration (mean (standard deviation) Dice = 0.92 (0.04) and Dice = 0.91 (0.05) (respectively) vs Dice = 0.80 (0.09), both p<0.001). Quantitative perfusion maps were also smoother, indicating a reduction of motion artifacts, with a median (inter-quartile range) standard deviation of 0.52 (0.39) ml/min/g in myocardial segments, than before motion correction and improved compared to the baseline method (0.55 (0.44) ml/min/g). Processing time was reduced by a factor of 15 for a representative image series using the deep learning approach in comparison to the iterative method.

**Conclusion**
The deep learning approach offers faster and more robust motion correction for stress perfusion CMR, improving accuracy for the dynamic contrast-enhanced data and the auxiliary images. It was trained with multi-vendor data and is not limited to a single acquisition sequence, so, as well as enhancing efficiency and performance, it could facilitate broader clinical use of quantitative perfusion CMR.

**Graphical abstract**

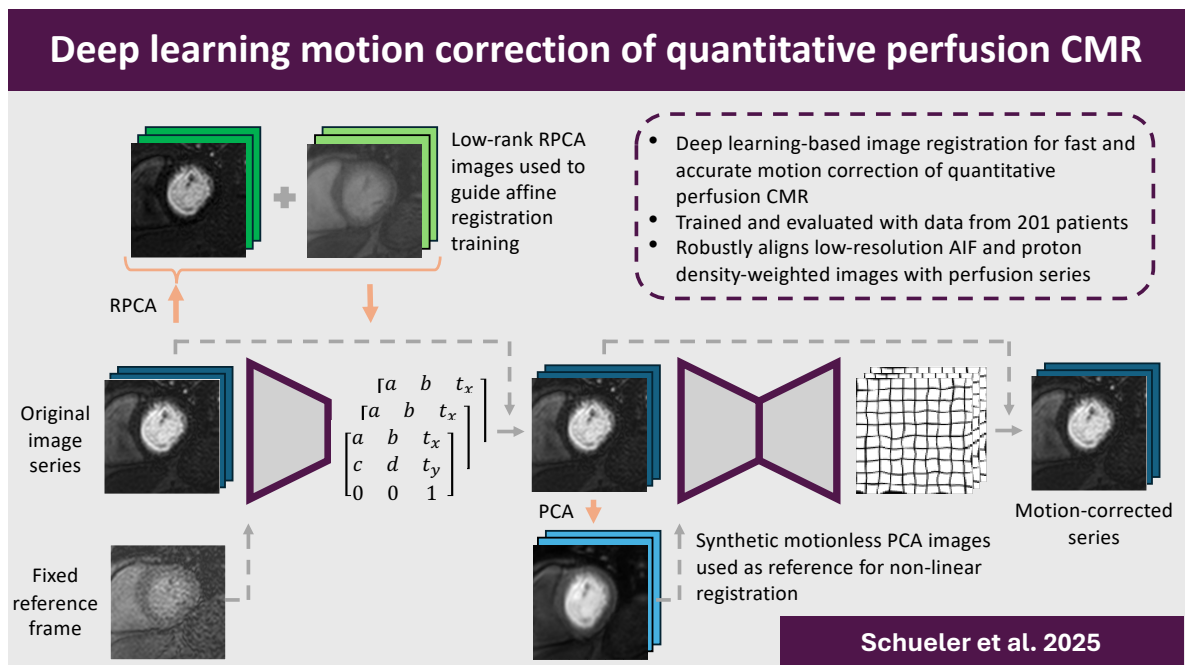

# List of abbreviations

| | |
|---|---|
| **AHA** | American Heart Association |
| **AIF** | arterial input function |
| **CMR** | cardiovascular magnetic resonance |
| **DSC** | Dice similarity coefficient |
| **ECG** | electrocardiogram |
| **IQR** | interquartile range |
| **LNCC** | local normalized cross-correlation |
| **LR** | low resolution |
| **LV** | left ventricle/left ventricular |
| **MONAI** | Medical open network for artificial intelligence |
| **NCC** | normalized cross-correlation |
| **PCA** | principal component analysis |
| **RPCA** | robust principal component analysis |
| **ROI** | region-of-interest |
| **SD** | standard deviation |
| **T** | Tesla |
| **TIC** | tissue intensity curve |

## Background

Stress perfusion cardiovascular magnetic resonance (CMR) is an established method for the assessment of myocardial ischemia with several indications in the clinical guidelines [1, 2]. However, accurate visual interpretation of the images is time-consuming and depends on the availability of highly experienced readers [3]. Quantitative myocardial perfusion analysis is, instead, user-independent, allowing objective assessment of the images and potentially wider clinical adoption of the modality. Objective thresholds of abnormal quantitative perfusion values have independent prognostic value [4, 5], and they can be easily used in less experience centers.

The evaluation of subendocardial ischemia requires high-resolution pixel-wise quantification to assess transmural gradients of perfusion [6, 7]. To achieve accurate and reproducible quantitative perfusion values at the pixel level, inter-frame misalignments need to be accounted for. While electrocardiogram (ECG) gating is used to account for cardiac motion, respiratory motion can be problematic. Due to the length of the acquisition (around 60 heartbeats), breath-holding is not sufficiently long to account for respiratory motion and for many patients no breath-holding is possible, so motion correction is required. Commonly, free-breathing acquisition protocols are employed with retrospective image-based motion correction using image registration. However, stress perfusion CMR visualizes a gadolinium-based contrast agent during its first pass through the myocardium. Therefore, dynamic contrast enhancement is taking place simultaneously to respiratory motion and represents a challenge for intensity-based image registration [8].

Several methods to motion correct stress perfusion CMR have been proposed [9–12]. Typically, these approaches preprocess the data, e.g. using principal component analysis (PCA), to circumvent the challenge of the dynamic contrast signal during registration. However, these approaches still rely on iterative optimization-based image registration algorithms which can lead to long computation times. As well as being iterative, several methods are progressive [13], i.e. a common approach gradually removes motion through several repetitions of PCA stages in which each stage involves an iterative registration. In a similar way, our previous work achieved good performance in a multi-stage approach by using both robust principal component analysis (RPCA), a matrix decomposition method, and PCA but is limited by computation time [14].

Recently, deep learning-based image registration approaches have gained popularity, as they significantly accelerate registration by estimating transformations in one shot, with both supervised and unsupervised image registration approaches being developed. Supervised methods involve training in which the ground truth transformations are known; and models are trained using a loss function based on the difference between the predicted and the ground truth transformation. However, ground truth deformations are usually not known and have thus been synthesized by applying known transformations to training data [15]. Conversely, unsupervised training applies the predicted transformation to the moving image and learns to minimize a loss function based on the dissimilarity between the fixed and moved images.

Previous work has shown that unsupervised deep learning image registration can capture the deformation of anatomical structures in cardiac motion estimation accurately [16]. For the similar problem of dynamic myocardial perfusion computed tomography motion correction, Lara-Hernández et al. also proposed an approach that included a recursive cascaded neural network, which implemented a loss function based on the Dice score between the left ventricle (LV) segmentation of the predicted and fixed frame as well as a contrast concentration loss [17]. Several applications have found that, in addition to being fast and accurate, deep learning registration was more robust than conventional optimization-based iterative registration algorithms [18–21]. This could be because deep learning models were exposed to a large representative sample of the expected transformations during training, and as a result, they learned to constrain predicted transformations to reasonable expected ranges based on the training data. On the other hand, iterative registration only considered the current image to be registered and usually estimated the required deformations from scratch for each new registration. Additionally for deep learning approaches, data of differing quality and appearances can be included during training, and perturbations of the input data can be simulated via data augmentation during training to improve robustness.

This robustness could be particularly beneficial for the motion correction of quantitative stress perfusion CMR data. Although algorithms typically address the dynamic-contrast enhancement during registration, there may still be residual contrast enhancement present during registration so robustness to variations in contrast is valuable. Additionally, accurate perfusion quantification necessitates the alignment of the proton density (PD) images, used for surface-coil intensity corrections, and the alignment of the low-resolution image series which is used to estimate the arterial input function (AIF) [22]. Both types of auxiliary images have a different appearance from the standard stress perfusion CMR data and may require specific tuning of registration parameters in optimization-based iterative algorithms, so robustness to these cases would be beneficial.

This work aimed to develop an unsupervised deep learning-based image registration approach to the motion correction of quantitative stress perfusion CMR data. A framework based on the previous approach of Scannell et al. [14] was designed but with the use of deep learning-based registration which enables fast one-shot registration. This approach does not require the RPCA preprocessing step to mitigate the effect of the contrast enhancement during registration, thereby further enhancing the efficiency of the motion correction. Additionally, it was hypothesized that AIF image series and PD images can be reliably corrected by the deep learning-based image registration approach. The deep learning models were trained with multi-vendor data from a varied patient cohort to further enhance the generalization ability of the framework, and they were validated using a range of quantitative metrics.

# Methods

**Data**

The complete retrospective dataset contained data of 201 patients who were clinically referred for the assessment of myocardial ischemia with stress perfusion CMR imaging at St Thomas' Hospital, London, UK. All patients provided written informed consent (regional ethics committee approval: 15/NS/0030) and were instructed to refrain from caffeine containing foods and drinks 24 hours before the scan. The CMR examinations in the development set were performed in free breathing using two different types of scanning systems. 157 patients were scanned using a 3-Tesla (T) Achieva TX system (Philips Healthcare, Best, The Netherlands) and were previously studied in Scannell et al. [23], whilst the other 44 patients were scanned using a 3T MAGNETOM Vida system (Siemens Healthineers AG, Erlangen, Germany) and were previously studied in Crawley et al. [24]. From this, an independent test set with data from 38 patients, scanned on the Philips 3T scanner was held-out to evaluate the registration performance. The remaining development set was further split into 149 patients for training and 14 for validation.

Perfusion image acquisition for both scanner vendors used previously described dual-sequence implementations with ECG-triggering [25, 26]. This included a low-resolution image series of the basal slice designed to minimize signal saturation in the AIF estimation. Each slice included two to three PD weighted images without saturation preparation, used for the surface coil intensity correction. High-resolution images were acquired in free-breathing for three short-axis slices covering the left ventricle (basal, mid, and apical) in addition to the low-resolution AIF slice, during adenosine-induced hyperemia (140–210 µg/kg/min, depending on the response to stress). The intravenous contrast agent was 0.075 mmol/kg of gadobutrol (Gadovist, Bayer, Berlin, Germany), injected at 4 mL/s, followed by a 25 mL saline flush at the same injection rate.

**Registration approach**

In this work, motion in stress perfusion CMR data was corrected in multiple stages, where bulk motion was corrected first using affine registration. To achieve this, RPCA was used to separate the contrast signal from the baseline (low rank) signal of the stress perfusion CMR data. The absence of dynamic contrast enhancement in the low rank image series allows easier registration to a common reference image and the estimated transformation can be applied to the original perfusion data to align the image series. The common reference image for the image series is the time dynamics that is ten dynamics before the end of the series. This was chosen as it was observed to exhibit a similar contrast appearance across different patients. After bulk motion correction, the next stages account for residual motion (affine and non-rigid) by registering the perfusion image series to synthetic motionless reference series created using PCA.

This approach, shown in Figure 1, was based on previous work of Scannell et al. [14], but replaced the optimization-based registrations with deep learning models that directly estimate the required transformations. Additionally, the first stage was modified so that RPCA is not required at runtime. That is, the first affine registration model was trained to estimate the affine transformation matrix directly from the original stress perfusion CMR image series. However, while not required to register new perfusion data, the low rank RPCA images were used during training to allow the deep learning model to learn to register without the confounding effect of the contrast enhancement. As shown in Figure 2, during training, as well as applying the predicted transformation to the moving perfusion CMR image to compute the loss with respect to the fixed perfusion CMR image, the predicted transformation was also applied to the corresponding low rank RPCA image so that loss function can also include the similarity between the fixed and moved RPCA baseline images. Since the transformation is predicted based on the perfusion CMR images only (shaded region in Figure 2), once trained, the RPCA images are no longer required. The deep learning registration models were trained in an unsupervised manner in that no ground-truth transformations or segmentations are used during training. The PD images preclude the separation of baseline signal and dynamic contrast-enhancement with RPCA so they have not been included in the previous method of Scannell et al. and cannot be used in the training of the deep learning approach. However, as RPCA is not required for inference with the deep learning approach the PD images can be included when the deep learning correction is run. As discussed in the Evaluation section, the capacity of the deep learning models to correct the PD images was tested, despite not being used in training.

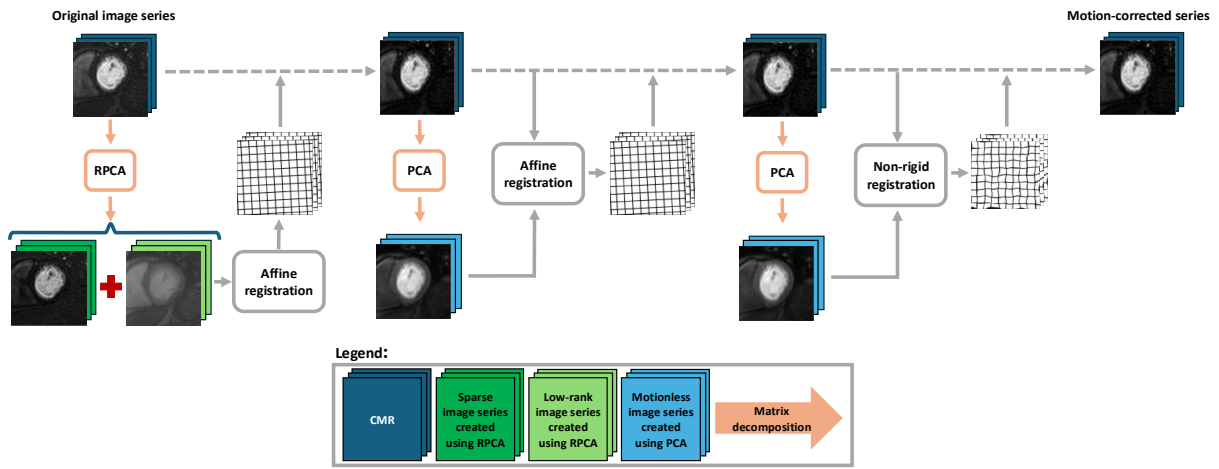

*Figure 1: Motion correction scheme based on Scannell et al. [14]. A three-step motion correction approach is used consisting of 2 affine image registration steps followed by a non-rigid image registration step. A combination of RPCA and PCA are used to mitigate the effect of contrast-enhancement and to create synthetic reference series for registration, respectively.*

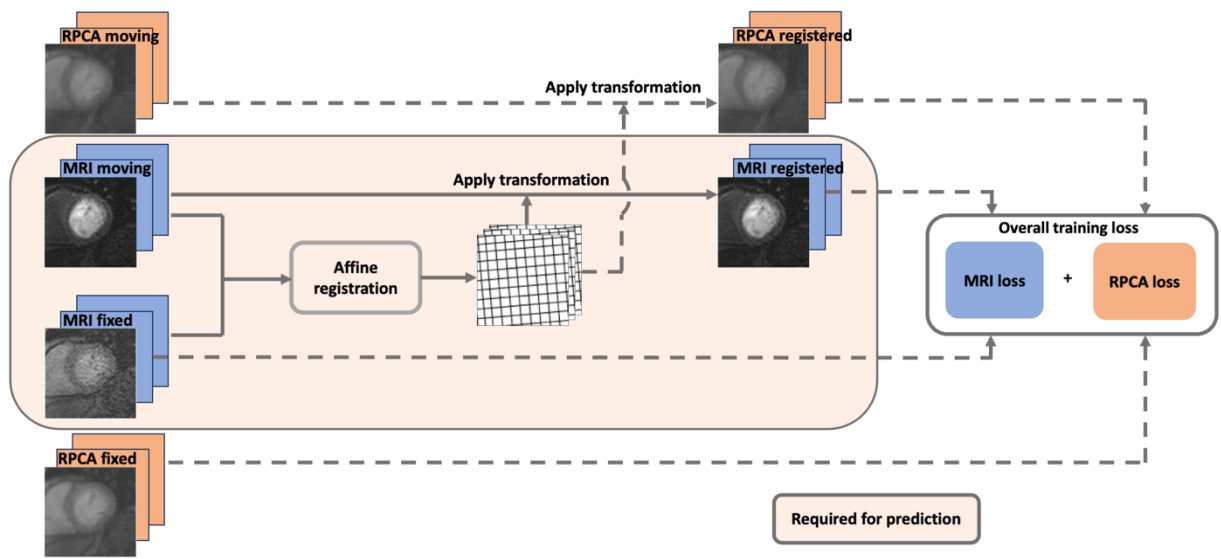

*Figure 2: The training and prediction approach of the first affine registration step from the overall pipeline. This approach predicts the affine transformation directly based on the stress perfusion CMR images. While training, the estimated transformations are also applied to the low-rank images extracted with RPCA and the total loss is a combination of the loss for the corrected perfusion series and the loss for the correct RPCA image series. Only the shaded region (not including RPCA images) is required at runtime. The MRI moving image is the time dynamic to be corrected and MRI fixed image is the reference image that the whole*

*series is corrected to. RPCA moving and fixed are the corresponding time dynamics from the low-rank RPCA series.*

**Model and training details**

Prior to training, perfusion images were preprocessed using a fully automated series of deep learning models, as described in Scannell et al. [27]. In summary, the image frame with the highest contrast signal in the LV was identified (peak LV). Based on this frame, a bounding box around the LV was determined to crop the image. The bounding box was adapted to a standardized size of 128x128 pixels for the affine registration steps and 96x96 for the non-rigid registration, and the cropped images were normalized to have intensity values in the range of 0-1. Histogram equalization was subsequently applied to the inputs of the deep learning models. PD images were not used during model training. Random augmentations were applied to the training images, to increase robustness and encourage the model to learn feasible transformations. During training of the affine models, rotation and translation transformations were applied to both moving and fixed images along with intensity augmentations. Intensity augmentations included intensity scaling and shifting of intensity values and noise addition. Larger affine augmentations were applied during the training of the first affine model compared to the second affine model, as it was expected that images entering the second affine model at run time were already better aligned. During training of the non-rigid model, only intensity augmentations were applied, as this model was intended to correct for fine misalignments. The specific hyper-parameters of the augmentations can be found in Supplementary Table 1.

The first affine model included 7 residual (ResNet) blocks, and the second affine model included 5 ResNet blocks [28]. The non-rigid model had a U-Net like architecture with an encoder depth of 4 ResNet down-sampling blocks followed by 4 up-sampling blocks [29]. Both the affine and non-rigid models had 16 initial channels, which were successively doubled, and the feature map sizes halved when down-sampling and (vice versa when up-sampling, in the case of the non-rigid model). In both affine models, a fully connected layer extracted the six affine transformation parameters from the last convolution layer for the 2-dimensional (D) affine transformation matrix. For the non-rigid model, a final convolutional output layer was used to predict dense displacement fields for the x and y direction.

The loss function for training the first affine registration step was the weighted sum of the negative normalized cross correlation (NCC) between the fixed and registered low-rank RPCA images (weight: 0.5), and the negative normalized mutual information between the RPCA images and CMR images (both combinations of fixed and registered images weighted with 0.25). The second affine registration step used the negative NCC loss function. The non-rigid registration model used the negative local NCC (LNCC), with a kernel size of 19, and was regularized by the bending energy penalty, which ensured smooth dense displacement fields by penalizing the second order derivative of the displacement field. The LNCC and bending energy were weighted at a ratio of 1:10 respectively. In this study, GlobalNet and LocalNet from the medical open network for AI (MONAI) framework [30] have been used to implement the affine and non-rigid registration networks [31]. The Adam optimizer was used to optimize all affine and non-rigid models with a fixed learning rate of 0.00001 and a batch size of 16 [32].

**Evaluation**

The proposed deep learning-based motion correction was evaluated against the existing iterative optimization-based solution of Scannell et al. [14] using a series of quantitative evaluation metrics on the held-out test set, and the run-time was compared for a representative test case. Additionally, an ablation study was performed to compare the proposed approach for predicting the required transformed in the first affine registration step using the perfusion CMR images versus an approach which used the RPCA low-rank images, similar to Scannell et al. [14].

Temporal smoothness within a circular region of interest (ROI) centered around the center of mass of the LV and expanded 10 pixels further than the widest point of the myocardium was evaluated. Temporal smoothness was measured based on the average standard deviation (SD) of the second order-derivative of the normalized time-intensity curves (TIC) of the pixels within the region. Based on the assumption that pixel values in well-aligned images within this region change in a smooth manner, solely due to the gradual passage of contrast over time, lower values of this metric indicate better registration. It was also assumed that a motion corrected image series will have high overlap of the myocardium between consecutive frames. Therefore, the Dice score between myocardium segmentations, obtained from a previously trained segmentation model [27], over five consecutive time frames around the peak LV time was evaluated.

PD images were not used to train the deep learning models but despite this they are still considered in the evaluation. To evaluate the capacity of the proposed approach to align PD images, the myocardium segmentation from the peak LV time dynamic is applied to the PD image to evaluate the standard deviation of pixel intensities within the segmented region. If the peak LV frame is well aligned with the PD image, the standard deviation of pixel intensities within the segmentation will be low, as it will include only myocardium and, thus, pixels should exhibit similar intensities.

Quantitative perfusion values were derived in the same manner as previous studies [33], using the dual sequence AIF, on a pixel-wise level using a Fermi function-constrained deconvolution [34]. The required image processing steps were achieved with a fully automated pipeline as previously described [27] and perfusion values were reported in 16 segments (defined by the American Heart Association (AHA) [35]). The standard deviation of pixel values across the 16 segments was reported and it was assumed that perfusion values in a well aligned series are more uniform across a segment than the values in a non-aligned series. To allow the visual assessment of registration quality, videos of the aligned image series are shown in the Supplemental Material.

Additionally, the runtime was compared between the deep learning approach and the original iterative registration for a representative test case. This evaluation compared the entire runtime including data loading and preprocessing steps: peak LV detection and bounding box calculation in addition to the motion correction. The comparison considered the same patient dataset and used the same standard hardware, a MacBook Pro M1 (Apple Inc., Cupertino, California) with 8-core CPU, without using a GPU.

**Statistical analysis**

Differences between the evaluation metrics for the different motion corrected approaches were analyzed using Wilcoxon signed-rank tests. The normality of these differences was assessed using the Shapiro-Wilk test.

# Results

**Study population**

The test data set baseline characteristics are summarized in Table 1.

*Table 1: Test patient characteristic (N = 38), values are n (%) or mean ± standard deviation (SD).*

| | |
|---|---|
| **Age (years)** | 61 ± 12 |
| **Male** | 24 (63) |
| **LV end-diastolic volume (ml/m$^2$)** | 86 ± 23 |
| **LV ejection fraction (%)** | 53 ± 14 |
| **Ischemia based on visual interpretation** | 21 (55) |
| **Presence of LGE** | 22 (58) |

**Temporal consistency**

The temporal smoothness of the image series, based on the median (inter-quartile range (IQR)) second-order derivative of the normalized TICs were 0.059 (0.01), 0.017 (0.007), and 0.015 (0.004) without motion correction (No MoCo), with the iterative optimization-based solution and with the proposed deep learning method, respectively. Our proposed deep learning method resulted in significantly smoother TICs compared to the iterative method, with p-values <0.001. The distribution of these values is shown in Figure 3. Note that a lower value of the mean 2nd derivative of the TIC indicates smoother TICs, and thus, better motion correction. It was found that the proposed deep learning method was improved in all slice locations, and in particular, it was seen that the performance for the proposed deep learning approach on the low-resolution AIF (LR-AIF) slices (right) was improved with respect to the iterative solution indicating the deep learning approach was better able to handle the motion correction of these image series.

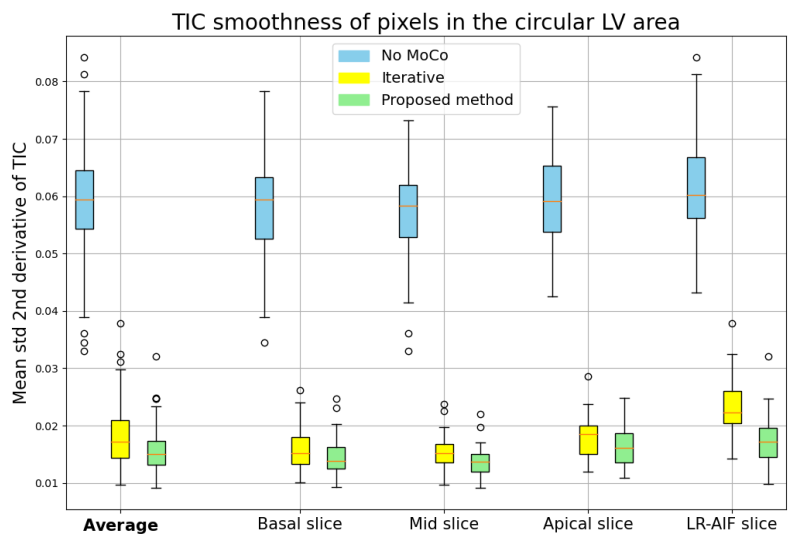

*Figure 3: A boxplot showing the distribution of TIC smoothness values comparing values before motion correction with the results after the iterative method and the proposed deep learning approach. Values are shown as an average over all slices and per individual slice. A lower value of the mean 2$^{nd}$ derivative of the TIC indicates smoother TICs. LR-AIF, low-resolution AIF images.*

The temporal overlap of the automated myocardium segmentation inferred on five consecutive time frames around the peak LV frame was also evaluated. The mean (SD) Dice similarity coefficient (DSC) over the test set demonstrated improved overlap from 0.80 (0.09) to 0.92 (0.04) after registration with the proposed deep learning method. The DSC was slightly improved compared to the original iterative optimization-based solution (0.91 (0.05)).

**PD images**

To assess the generalization of the proposed model to the PD images, the alignment of the automated myocardium segmentation from the perfusion series to the structure of the PD images was considered. Despite not being used in training the deep learning method, the median (IQR) standard deviation of PD pixels within the myocardium segmentation for the proposed deep learning method was 0.10 (0.05), a significantly lower standard deviation than prior to motion correction 0.15 (0.08), p-value < 0.001. This indicates that the PD images were better aligned to the perfusion image series after motion correction with the proposed method. An overlay of the PD image with a time frame of the perfusion image series for a representative test case is shown in Figure 4, further illustrating the improvement of the alignment after motion correction. Comparison to the original solution was not performed as PD images were not included in the method of Scannell et al. [14], as they were not available at the time of development.

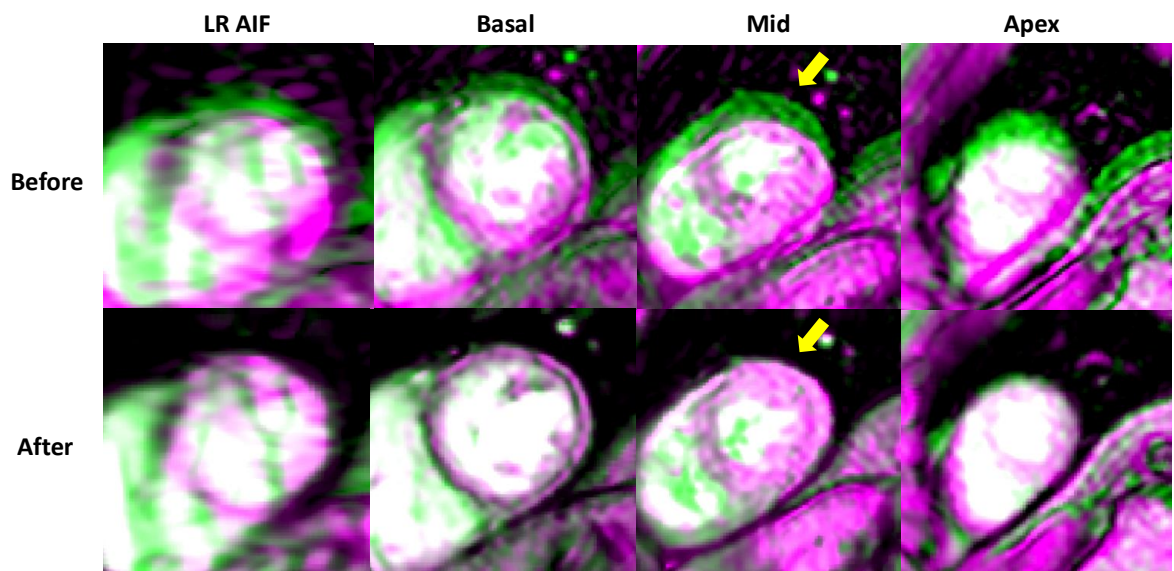

*Figure 4: Color channel overlay of PD image (green) with a time frame from the perfusion image series (red) before (top) and after motion correction (bottom) with our proposed method, for the low-resolution AIF slice (LR AIF) and three high resolution slices (left to right). The arrow highlights an area of misalignment before motion correction that is noticeably improved after motion correction.*

**Quantitative perfusion**

The median (IQR) quantitative perfusion value without motion correction was 3.59 (2.56) ml/min/g, compared with 2.19 (1.15) ml/min/g after iterative registration and 2.17 (1.14) ml/min/g after the proposed deep learning-based motion correction. The high values without motion correction indicates the presence of motion artifacts which manifest as high perfusion values. Figure 5 shows quantitative perfusion maps, for a patient assessed as having no visible perfusion defects, after motion correction. The relative homogenous appearance of the perfusion values is indicative of a lack of motion artifacts in the image series. Additionally, the smoothness of perfusion maps is quantified as the standard deviation of perfusion values within the AHA segments. The median (IQR) standard deviation of perfusion in AHA segments after motion correction is 0.52 (0.39) which is significantly improved compared to before motion correction of 1.02 (0.75), p-value < 0.001, and lower than with the iterative registration approach; 0.55 (0.44), p-value < 0.001.

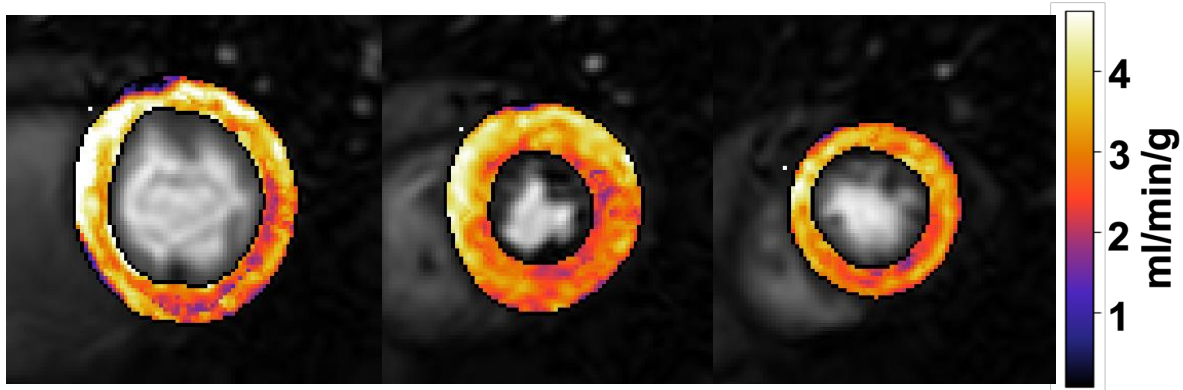

*Figure 5: An example case from the test data set with the quantitative perfusion maps showing relatively homogenous perfusion values and a lack of motion artifacts.*

To allow for visual assessment, videos showing before and after motion correction for the image series of three representative test cases, each including three high-resolution slices and the low-resolution AIF slice with PD images, are included in the Supplementary material.

**Runtime**

The runtime for a representative test case, with four slices (the LR-AIF slice and three standard perfusion slices) and 71 acquired time dynamics, including data loading and preprocessing steps: peak LV detection and bounding box calculation in addition to the motion correction was 47.9 seconds, on standard hardware without using a GPU. Motion correction on each slice took 7.6 seconds on average. This is over 15 times faster than the runtime of the original iterative optimization-based solution, which was 12 minutes 32 seconds for this case without including the preprocessing steps (direct comparisons cannot be made due to implementation differences).

**Ablation study**

An ablation study was performed to investigate the effect of using the perfusion CMR images directly as input to the first deep learning registration, as shown in Figure 2, instead of using the low-rank RPCA images as was proposed in Scannell et al. [14]. The RPCA input method resulted in similar median TIC smoothness values of 0.015 (0.004), mean Dice overlap of 0.92 (0.04), median standard deviation of myocardium pixels in the PD images of 0.10 (0.05), and median standard deviation of perfusion in AHA segments of 0.52 (0.39). None of these values were improved compared to the proposed method which suggests that the use of RPCA at runtime is not necessary. Additionally, due to the need to compute the low-rank RPCA images at runtime, this approach takes more than double the time of our proposed method (104.1 seconds).

# Discussion

In this work, we developed and evaluated a fast deep learning image registration approach for free-breathing quantitative stress perfusion CMR. A strength of this work is that the deep learning models were trained with a large multi-vendor dataset, and the deep learning registration pipeline outperforms a previously described iterative registration pipeline with respect to several quantitative metrics as well as substantially reducing the required computational time. The deep learning method significantly outperformed the iterative methods when aligning low resolution AIF image series and it also well-aligned the PD images. This showed its capacity to handle variations in the input data well and generalize to image types that were not used during training. The motion correction is not dependent on the image acquisition and is not tied to a scanner vendor or particular acquisition method. The flexibility of this approach can potentially contribute towards more widespread of quantitative stress perfusion CMR.

The quantification of the smoothness of myocardial time-intensity curves showed a significant improvement for the deep learning method over the previous iterative registration solution, and the analysis of the temporal consistency of myocardium segmentations and smoothness of quantitative perfusion maps confirmed the strong performance of the deep learning registration method. The effect of bypassing the RPCA calculation was evaluated by training an additional deep learning model that corrects the first affine motion based on low-rank RPCA images. Both methods had a similar performance when evaluating the registration, but the proposed method is substantially faster as it does not require the RPCA calculation at runtime.

The motion correction of contrast-enhanced perfusion CMR data with intensity-based registration algorithms is challenging due to the effect of the dynamic contrast changes. Several prior works in the field tackled this problem but all of these employed iterative optimization-based image registration algorithms [9, 11, 12, 14]. Our work is the first to approach this with deep learning registration, and our results showed several advantages of deep learning registration in this application. Particularly, since the deep learning registration models were trained to predict transformations directly from the fixed and moving input images, the traditional iterative optimization of a similarity metric is circumvented. Computational efficiency is important as the field moves towards near real-time processing of the data at scan time. Also, since optimization of a similarity metric is not required when the model is applied, the challenge of computing the similarity of images at different stages of dynamic contrast enhancement is mitigated. Instead, the deep learning models learned during training to deal with images of varying contrast and can directly align the images in a fast and robust manner.

It is also interesting that the deep learning model architectures used were based closely on those used previously for prostate MRI data [31], using the implementations taken directly from the MONAI framework [30], without the need for significant adaptations or fine-tuning of hyperparameters. This indicates that this approach generally performs well irrespective of the specific data and application. Our models are also likely to generalize well to different patient groups as the training data consisted of an unselected population representative of those seen in clinical practice, and as shown in Table 1, the models were tested in a varied cohort including a high proportion of patients with varying levels of disease.

Going forward, the deep learning motion correction pipeline may have further benefits because methods that combine motion correction with the image reconstruction [36], or the tracer-kinetic modelling [37] are being actively explored, and both topics are being increasingly addressed with deep learning methods [38, 39], potentially leading to synergy with our deep learning motion correction.

**Limitations**

While the proposed method did demonstrate good generalization capabilities, as evidenced by its accurate alignment of PD images, low-resolution AIF image series, and standard high-resolutions image series with the same models, further investigation of the robustness to varying clinical scenarios (different hospitals, scanners, acquisition protocols, etc.) is warranted. The models were trained with data from two different MRI scanner vendors, however, since the test set was limited to data from a single scanner from a single hospital, follow-up testing should be performed with data from additional centers.

The training data did include some examples of cardiac motion (e.g. caused by mistriggering during acquisition) and through-plane motion. However, the focus of this work was specifically to correct respiratory motion, and the amount of such cases was small. Visual inspection indicated that large amounts of cardiac motion is still not always corrected well, and future work will look into recognizing these frames and developing algorithms to specifically correct cardiac motion or excluding them from the perfusion quantification. Motion correction and evaluation were performed only in 2D with motion in the third dimension not considered, as is typical for perfusion CMR, due to the large slice thickness and slice gap. As methods are developed to acquire data in 3D [40] or to increase the number of acquired slices [41], 3D registration will also need to be considered.

Also, as seen by the temporal smoothness metric in Figure 3 and the visual assessment of the motion corrected series, some residual motion remained in the apical slices after motion correction. This can be attributed to the reduced thickness of the myocardium in this slice and the potentially more complex motion patterns (e.g. through-plane motion) being more difficult to correct. To address these remaining challenges, future work could consider adding temporal information to the model input by considering the whole image series to be aligned rather than correcting in a frame-by-frame manner. Such an approach circumvents the reference frame selection bias, provides the model with more information about the underlying motion pattern in the images series and has been shown to work well for cardiac $T_1$ mapping [42].

Finally, registration evaluation is a difficult topic in general [43], and there was no ground-truth possible for validation in this work. To account for this, we quantified several different quantitative metrics, each of which give an indication of an aspect of the motion correction performance, but the ultimate validation would compare measurements of ischemia given by the resulting quantitative perfusion maps versus an objective reference standard such as fractional flow reserve or alternative imaging modalities.

## Conclusion

This work developed a deep learning registration pipeline, trained on a varied multi-vendor dataset, for the motion correction of stress perfusion CMR data, with the goal of improving the previous iterative optimization-based solution while also improving the time efficiency. The unsupervised registration pipeline, which combines two affine models and a non-rigid model, achieves this goal. This pipeline significantly improved performance on two evaluation criteria, matched the iterative solution on the other criterion, and is substantially faster than the iterative method. Importantly for perfusion quantification, it also performs well for the motion correction of the low-resolution AIF slice and well aligns the PD-weighted images with the dynamic contrast-enhanced perfusion data.

# Funding

The authors acknowledge financial support from the Department of Health (DoH) through the National Institute for Health Research (NIHR) comprehensive Biomedical Research Centre award to Guy's & St Thomas' NHS Foundation Trust in partnership with King's College London and King's College Hospital NHS Foundation Trust and by the NIHR MedTech Cooperative for Cardiovascular Disease at Guy's and St Thomas' NHS Foundation Trust. This research was also funded in part by the Wellcome Trust via the joint Wellcome Trust / Engineering and Physical Sciences Research Council (EPSRC) Centre for Medical Engineering award [203148/Z/16/Z] and the Wellcome Trust Innovator Award [222678/Z/21/Z]. For the purpose of open access, the author has applied a CC BY public copyright licence to any Author Accepted Manuscript version arising from this submission

# Supplementary material

**Data augmentation**

Data augmentations were applied during training of the affine and non-rigid registration models. Hyper-parameters of the intensity augmentations were the same during all training and are shown in Supplementary table 1. Gaussian noise with a zero mean and standard deviation of 0.01 is always applied. Besides, the intensity was scaled by a random factor uniformly sampled in the range of -0.3 to 0.3 and shifted by a random offset uniformly sampled in the range -0.2 to 0.2.

*Supplementary table 1: Hyper-parameters of intensity augmentations, applied to the first and second affine and non-rigid model of both approaches.*

| Augmentations | Probability | Hyper-parameter 1 | Hyper-parameter 2 |
|---|---|---|---|
| Gaussian noise | 1 | Mean: 0 | Standard deviation: 0.01 |
| Scale intensity | 1 | Lower: -0.3 | Upper: 0.3 |
| Shift intensity | 1 | Lower: -0.2 | Upper: 0.2 |

Geometric augmentations were applied during training of the first and second affine registration models. Larger affine augmentations were applied to the first affine model in the pipeline, relative to the affine augmentations applied to the second affine model in the pipeline as can be seen in Supplementary table 2, to encourage the second model to focus on smaller translations and rotations. A random translation range, which is a translation in number of pixels, and a random rotation range, which is an angle in radians, is uniformly sampled from the given ranges for every image during training.

*Supplementary table 2: Hyper-parameters of rigid augmentations, different for the first and second affine model.*

| Approach | Probability | Translation range (nr pixels) | Rotation range (angle in radians) |
|---|---|---|---|
| First affine model | 1 | (-20, 20) | (-0.8, 0.8) |
| Second affine model | 0.5 | (-10, 10) | (-0.4, 0.4) |